\pdfoutput=1

\documentclass[11pt]{article}

\usepackage[]{acl}
\usepackage{tabularx}
\usepackage{multirow}
\usepackage{stfloats}
\usepackage[labelfont=bf]{caption}
\usepackage{times}
\usepackage{latexsym}

\usepackage[T1]{fontenc}

\usepackage[utf8]{inputenc}

\usepackage{microtype}

\usepackage{graphicx}
\usepackage{graphics}

\title{Searching for Snippets of Open-Domain Dialogue in Task-Oriented Dialogue Datasets}

\author{Armand Stricker,   Patrick Paroubek \\
  Université Paris-Saclay, CNRS, \\
  Laboratoire Interdisciplinaire des Sciences du Numérique, \\
  91400, Orsay, France \\
  \texttt{firstname.lastname@lisn.upsaclay.fr}
  }

\begin{document}
\maketitle
\begin{abstract}
Most existing dialogue corpora and models have been designed to fit into 2 predominant categories : task-oriented dialogues portray functional goals, such as making a restaurant reservation or booking a plane ticket, while chit-chat/open-domain dialogues focus on holding a socially engaging talk with a user.  However, humans tend to seamlessly switch between modes and even use chitchat to enhance task-oriented conversations.  To bridge this gap, new datasets have recently been created, blending both communication modes into conversation examples.  The approaches used tend to rely on adding chit-chat snippets to pre-existing, human-generated task-oriented datasets.  Given the tendencies observed in humans, we wonder however if the latter do not \textit{already} hold chit-chat sequences.  By using topic modeling and searching for topics which are most similar to a set of keywords related to social talk, we explore the training sets of Schema-Guided Dialogues and MultiWOZ. Our study shows that sequences related to social talk are indeed naturally present, motivating further research on ways chitchat is combined into task-oriented dialogues.   
\end{abstract}

\section{Introduction}

Over the past few years, building dialogue systems that converse with humans in a natural and authentic way has gained in popularity \citep{Ni2021}.  Although the ultimate goal is to build a single system capable of human communication in all of its intricacy and complexity, models generally fall into 2 categories : task-oriented dialogue (TOD) agents (\citealp{Hosseini2020SimpleTOD}; \citealp{Peng2020}; \citealp{ham-etal-2020-end}) and chit-chat or open-domain dialogue (ODD) agents (\citealp{Adiwardana2020}; \citealp{roller2020}; \citealp{zhang2019dialogpt}).

TOD systems use dialogue to help users complete tasks, such as airline or restaurant booking \citep{bordes2016}, and research on these systems typically aims to improve task success rate—meeting the users’ goals—in as few turns as possible \citep{deriu2021}.  In contrast, ODD systems are designed for extended conversations, mimicking the unstructured exchanges characteristic of human-human interaction \citep{roller2020recipes}.  
However, especially when considering a task-oriented setting, blending these 2 communication skills appears to be a desirable trait for several reasons.  

Firstly, building rapport and common ground through small talk/chit-chat or more generally, ODD, can be crucial to the establishment and maintenance of any collaborative relationship, an essential aspect of task-oriented dialogues \citep{bickmore2001}.  Small talk can help in building trust and in easing cooperation, by ‘greasing the wheels’ of task talk.  In fact, engaging in social talk has been associated with better task outcomes in several domains (\citealp{Wentzel1997StudentMI}, \citealp{levinson1999}).  

Secondly, when performing tasks through conversation or another mode, people tend to have not one but multiple possibly underlying goals \citep{Reeves1996TheME} : blowing off steam, impressing one’s husband or wife, avoiding boredom…. A conversation is situated, and peripheral information tends to seep into the conversation, even when the goal of the latter seems quite explicit. 

Numerous datasets have been created for TOD and ODD in recent years, however only very few of them take into account the possible overlap between task-oriented and open-domain conversation in human-human dialogue.  Recent efforts have been made to fill this void, by augmenting human-generated TOD datasets with chit-chat.  Accentor \citep{Accentor2020} propose to decorate system responses in the Schema-Guided Dialog (SGD) dataset \citep{sgdrastogi2019} with ODD snippets, making the dialogue agent sound more engaging and interactive.  FusedChat \citep{Fusechat-young-2021} appends and prepends human-written ODD to TODs from the MultiWOZ dataset \citep{budzianowski-etal-2018-multiwoz} and focuses on transitioning from one type of dialogue to the other, treating TOD and ODD as parallel dialogue modes.  Both of these approaches assume that the utterances in the task-oriented datasets are all strictly task-related and that chit-chat must be \textit{added} for it to be present in the dialogue.  
Although this seems reasonable to expect, since SGD and MultiWOZ’s collection guidelines are strictly task-related, we wonder,  given the reasons stated previously, if instances of ODD are not \textit{already} present in the task-oriented conversations.  

Topic modeling has been shown to help discover new content via corpus exploration \citep{mimno-mccallum-2007} and we therefore sift through the training sets of SGD and MultiWOZ, searching for topics which are most similar to a set of ODD-related keywords.  We find that certain sequences from the datasets \textit{are} related to ODD.  This suggests that social talk and task-oriented dialogue are indeed naturally intertwined, and that this aspect should be taken into account when building TOD datasets in the future, to more closely recreate  natural dialogues systems can learn from.

\section{Data}

\subsection{MultiWOZ}
MultiWOZ is a task-oriented dataset with a training set of more than 8,000 dialogues spanning multiple domains such as bus and taxi reservation, restaurant and train booking…  The data collection is done by using the Wizard of Oz framework \citep{Kelley1984AnID}, where a human user  unknowingly interacts with another human, who plays the role of the system.  In this particular instance, each task is mapped to a natural language description, to guide the user.  In principle, this leaves little room for off-script utterances that do not move the task forward.  Furthermore, to ensure data quality, a separate group of crowd-workers hired to annotate the data with dialog acts, can report errors when the dialog does not follow the task guidelines or if confusing utterances are present.

\subsection{The Schema-Guided Dataset}
SGD also covers a wide variety of domains and has more than 16,000 dialogues. It also introduces a novel data collection approach : the authors develop a multi-domain dialogue simulator that generates dialogue skeletons. The simulated agents interact with each other using a finite set of actions which are then converted into natural language utterances using a set of templates.   Humans are added to the loop only to paraphrase the templatized utterances and make the dialogues more natural and coherent.  This data collection method also leaves little room for open-domain sequences of text.

\section{Approach}
\subsection{Topic Model} 
We choose to use BERTopic\footnote{\url{https://github.com/MaartenGr/BERTopic}} \citep{bertopic}, a state-of-the-art topic model which generates topics in three steps.  First, each document is converted to its embedding representation using a pre-trained language model :  we use the 'all-mpnet-base-v2' model from the Sentence-BERT (SBERT) framework\footnote{\url{https://www.sbert.net/}} \citep{sbert-reimers} as it achieves state-of-the-art performances on several embedding tasks (\citealt{reimers-gurevych-2020-making}; \citealt{thakur-etal-2021-augmented}). Then, using UMAP \citep{mcinnes-umap} the dimensionality of these embeddings is reduced to optimize the next step, the clustering process, which is done using the HDBSCAN algorithm \citep{mcinnes-hdbscan}.  Finally, topic representations are extracted from the clusters using a class-based variation of TF-IDF : all documents in a cluster are treated as a single document and c-TF-IDF\footnote{\url{https://github.com/MaartenGr/cTFIDF}} then models the importance of words within the clusters, generating topic-word distributions for each cluster of documents.  As an extra step, the number of topics is reduced by iteratively merging the c-TF-IDF representations if the similarity scores between topic embeddings exceeds 0.915. This threshold is the one implemented by default in the framework.

\subsection{Set of ODD-related Terms}
Once the topics are generated, we find those most similar to a set of key words/phrases related to ODD. One single topic is represented by a list of up to 10 words, and using the same SBERT model mentioned previously, we embed a topic by averaging the embeddings of each word in the list.  We then embed each ODD-related keyword and compute a similarity score.  Inspired by the categories created by \citet{dunbar1997human} to classify conversations observed in informal social situations, we propose to experiment with the following key words as ODD references : \textit{personal relationships; personal experiences; emotional experiences and feelings; sport and leisure; work and school}.

\subsection{Inputs}
For each dataset, we experiment with two different inputs for the topic modeling algorithm: the full raw utterances and a set of filtered clauses extracted from the same utterances.  

We inspect the utterances at this finer-grained level given the implemented safeguards against ODD in the data collection process, similarly to other tasks which require fine-grained textual analysis \citep{gui-etal-2016-event}.  Clauses in English grammar are defined as the smallest grammatical structures that contain a subject and a predicate, and can express a complete proposition \citep{kroeger2005analyzing}.  Indeed, this segmentation produces more documents for BERTopic to analyze with less, more condensed information: this helps detach potential ODD snippets from their TOD contexts. 

To split utterances into clauses, we apply \citet{oberlander-klinger-2020-token} clause extraction algorithm, designed to separate text into clauses for emotion stimulus detection, and which achieves an F1 score of up to 80\% for clause detection.  For example, the utterance “{\it Find me a comedy to watch right now.  I’m super bored.}” is split into the following list of clauses
{\fontfamily{qcr}\selectfont
["Find me a comedy", "to watch right now", "I'm super bored"]
}.

Furthermore, because we have access to the datasets’ annotations, we filter out clauses which may contain task-related information through several steps.  For each utterance, we create a string which concatenates domain, intent, slot and value information. For the example above, this will yield  "\textit{movie, play movie, genre comedy}”.  If there is any overlap between these tokens and tokens in the clause (stopwords and punctuation excluded), we consider the clause to be task-related. If the clause detection algorithm detects only a single clause in the utterance, we also consider the clause to be task-related, since every utterance is supposed to help move closer to the task goal.  Finally, using SBERT, we keep only the clauses whose embeddings are least similar to the embeddings of the concatenated strings. This returns the string "{\it I'm super bored}" in the previous example.    


\section{Results and Discussion}
We retrieve the topics most similar to each keyword, looking at each model (2 models per dataset, due to the different versions of the inputs considered).  We look at the top 5 topics in each case and explore the sequences that the model has assigned to these topics.  Our findings show that our approach is most successful for SGD, with the filtered clauses as input.  Indeed, we find open-domain sequences for all of the keywords.  These include mentions of going on a date for "\textit{personal relationships}", having a great time for "\textit{personal experiences}", feeling sick and unwell for "\textit{emotional experiences and feelings}", being bored and having nothing to do for "\textit{sports and leisure}", and having a vacation coming up for "\textit{work and school}".  We report the relevant topics (defined by up to 10 topic words) with which the sequences are associated in table \ref{table:1}.  As we can see, certain topics come up multiple times, possibly due to semantic overlap in the keywords. We also report sequence examples in appendix \ref{sec:appendix}.

\begin{table*}[h]
\begin{tabular}{|l|l|}
\hline
\multicolumn{1}{|c|}{ODD-related Keywords}                                                                & \multicolumn{1}{c|}{Topic words}                                                                     \\ \hline
\multirow{2}{*}{Personal relationships}                                                       & \textit{meeting, friend, dining, dinner, friends, client, later, date, old, tonight}            \\ \cline{2-2} 
                                                                                              & \textit{friend, friends, husband, willing, my, only, person, best, and, we}                     \\ \hline
Personal experiences                                                                          & \textit{experience, time, fun, have, great, good}                                               \\ \hline
\multirow{2}{*}{\begin{tabular}[c]{@{}l@{}}Emotional experiences\\ and feelings\end{tabular}} & \textit{feeling, sick, feel, unwell, better, lately, well, cabin, earache, dizziness}           \\ \cline{2-2} 
                                                                                              & \textit{experience, time, fun, have, great, good}                                               \\ \hline
Sport and leisure                                                                             & \textit{office, relaxing, winding, refreshment, routine, routing, clue, jobs, intesting, among} \\ \hline
\multirow{2}{*}{Work and school}                                                              & \textit{vacation, holidays, soon, school, unused, vacationing, our, coming, taking, family}     \\ \cline{2-2} 
                                                                                              & \textit{office, relaxing, winding, refreshment, routine, routing, clue, jobs, intesting, among} \\ \hline
\end{tabular}
\caption{\label{table:1} To extract potential ODD sequences, we search for the topics which are most similar to a set of hand-picked ODD-related keywords.  In this table, we report the topics (defined by 10 topic words at most) that are associated with sequences we found to be relevant in the filtered version of SGD.}
\end{table*}
The rest of the associated sequences seem to be largely made up of clauses without enough context ("\textit{to live ?}", "\textit{is this ?}", "\textit{I'm planning on}") or clauses strongly associated to TOD ("\textit{Business or economy is fine}", "\textit{and Fighting with My Family are all playing}").  Splitting the utterances into clauses can create noisy, very short sequences that are hard to associate with either TOD or ODD, which explains the first scenario.  The second scenario can be explained by the fact that the annotations can sometimes be incomplete. "\textit{Fighting with My Family}" is a movie at the end of a list that the system offers to the user.  However, the annotation only mentions the first movie in that list.  As for "\textit{Business or economy is fine}", this illustrate the limits of our filtering approach. Although keeping only the clause that is least similar to the annotations helps in separating out sequences with task-oriented intents, this task is a difficult one and would require more precision in the approach.  

The unfiltered version of SGD mainly contains task-related utterances.  Some examples however, associated with the keyword "\textit{personal relationships}", illustrate how ODD and TOD can be intertwined in the same utterance : "\textit{I'm taking a friend out to dinner. Can you recommend a place to eat nearby?}" or "\textit{I almost forgot I have a date coming up I need to plan for! Can you look up some places to eat for me?}".  Examples such as these and the fact that more sequences were extracted from the filtered than the unfiltered version support the idea that humans, acting as dataset annotators in this context, tend to naturally \textit{intertwine} ODD and TOD in SGD, even in the absence of explicit instructions to do so.

As for MultiWOZ, the unfiltered version yields only task-oriented utterances and the filtered version produces clauses which would need more context, or clauses identifiable as TOD.  We invoke the same reasons mentioned previously, as well as the fact that the MultiWOZ data collection process is heavily guided and that the utterances are "proof-read" by a different set of annotators, which is not the case for SGD.  The only relevant mentions are linked to the "\textit{personal experience}" keyword and include sequences such as "\textit{I'm so bored, can you help me}", "\textit{That sounds like fun !}" or "\textit{Enjoy your time}", which nevertheless shows that ODD does exist in MultiWOZ, albeit in lower proportions.

\begin{figure}[htbp]
\centerline{\includegraphics[scale=.5]{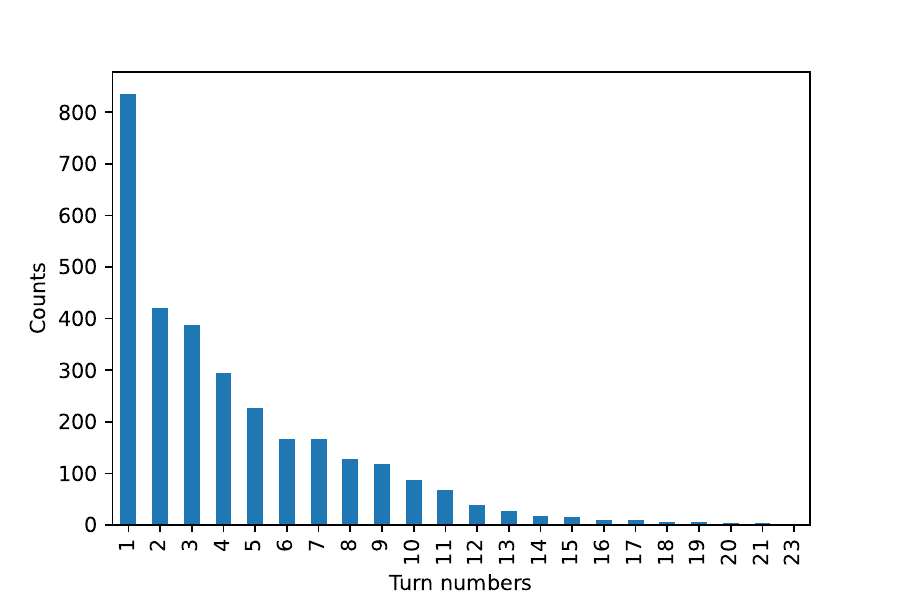}}
\caption{Distribution of turn numbers for all of the sequences extracted from the filtered version of SGD}
\label{fig:distrib}
\end{figure}

Furthermore, our results give an indication of how a single utterance can combine ODD and TOD and what purpose this may serve.  We find that many extracted sequences illustrate the fact that ODD can allow a user to justify their request, adding personal details about how or why they need the system's services, as in "\textit{I'm going on a date and want to take her to dinner. Can you look up places to eat?}" . This is further indicated by figure (\ref{fig:distrib}), which shows the distribution over dialogue turns of the ODD sequences extracted from the filtered SGD corpus. We observe that most sequences are bound to the first turn of the dialogues, which is where one might mainly expect to see some form of justification, as a form of collaborative self-disclosure before engaging in TOD.


\section{Conclusion}
Sifting through the training sets of SGD and MultiWOZ, we find that ODD sequences do exist, especially in SGD, and that these modes of communication do not necessarily need to be treated separately.  Both modes can be present within a single utterance and future task-oriented datasets may amplify this observation, in an effort to propose more natural TOD resources for models to learn from.

\bibliography{Prospecting_SHORT}

\appendix
\onecolumn
\section{Sequence Examples}

\label{sec:appendix}
We show, for each model, a sample of the sequences extracted, along with turn number and speaker information. 

\begin{table*}[!h]
    \centering
    \begin{tabular}{|l|r|p{.4\linewidth}|r|c|}
    \hline
        ODD-related Keywords & Topic & Sequence Examples & Turn & Speaker \\ \hline \hline
        personal relationships & 899 & and Fighting with My Family . & 2 & system \\ \hline
        personal relationships & 342 & I have a date night coming up & 1 & user \\ \hline
        personal relationships & 1094 & My husband and I want & 1 & user \\ \hline
        personal relationships & 109 & What are the other dates ? & 5 & user \\ \hline
        personal relationships & 1054 & I 'm fine with either business & 4 & user \\ \hline
        personal experiences & 1231 & , have a fun experience ! & 5 & system \\ \hline
        personal experiences & 876 & Do I have the details correct ? & 4 & system \\ \hline
        personal experiences & 395 & they have bad reviews & 4 & user \\ \hline
        personal experiences & 1068 & Details ? & 1 & system \\ \hline
        personal experiences & 684 & I like . & 10 & user \\ \hline
        emotional experiences  & 714 & How are you doing today ? & 8      & user \\ 
          and feelings         &   \hfill  &        \hfill       & \hfill &     \hfill     \\ \hline
        emotional experiences  & 185 & I am in a nice mood & 1 & user \\ 
          and feelings         &   \hfill  &        \hfill       & \hfill &     \hfill     \\ \hline
        emotional experiences & 493	& I 've been feeling sort of unwell recently & 1 & user \\ 
          and feelings         &   \hfill  &        \hfill       & \hfill &     \hfill     \\ \hline
        emotional experiences  & 1231 & have a good time ! & 5 & system \\ 
          and feelings         &   \hfill  &        \hfill       & \hfill &     \hfill     \\ \hline
        emotional experiences  & 558 & Could you tell me how to & 12 & user \\ 
          and feelings         &   \hfill  &        \hfill       & \hfill &     \hfill     \\ \hline
        sport and leisure & 426 & Sports ? & 1 & system \\ \hline
        sport and leisure & 652 & where you want the attractions to be ? & 1 & system \\ \hline
        sport and leisure & 374 & I would like to look for games & 4 & user \\ \hline
        sport and leisure & 1087 & and I like games & 8 & user \\ \hline
        sport and leisure & 914 & I can use some time away from the internet . & 1 & user \\ \hline
        work and school & 719 & I can not wait to go on a vacation . & 1 & user \\ \hline
        work and school & 914 & I feel like winding down and relaxing . & 1 & user \\ \hline
        work and school & 866 & to unwind . & 1 & user \\ \hline
        work and school & 46 & Would that work ? & 3 & system \\ \hline
        work and school & 615 & that would work for your timeframe . & 2 & system \\ \hline
    \end{tabular}
\caption{Example sequences from the \textbf{filtered SGD} topic model}
\end{table*}

\begin{table*}[b]
    \centering
    \scalebox{0.79}{%
    \begin{tabular}{|l|r|p{.6\linewidth}|r|c|}
    \hline
        ODD-related Keywords & Topic & Sequence Examples & Turn  & Speakers \\ \hline \hline
        personal relationships & 1797 & Family sounds good & 4 & user \\ \hline
        personal relationships & 1489 & Anything else in economy? & 11 & user \\ \hline
        personal relationships & 822 & I'm hanging out with my friend later and need to find somewhere for us to eat. Can you look up restaurants for me? & 1 & user \\ \hline
        personal relationships & 672 & \$15 per head and this is a drama movie. & 11 & system \\ \hline
        personal relationships & 2437 & What airport would I have to depart from for this flight? Oh yeah. And tell me if this an economy or business class ticket? & 11 & user \\ \hline
        personal experiences & 3011 & Tell me more about your personal tastes & 9 & system \\ \hline
        personal experiences & 457 & Are there any Life history movies? & 8 & user \\ \hline
        personal experiences & 877 & Which date would you like to know? & 5 & system \\ \hline
        personal experiences & 1930 & I want to book a house and will check out on the 14th. & 1 & user \\ \hline
        personal experiences & 733 & Does that restaurant have live music? & 14 & user \\ \hline
        emotional experiences & 2723 & I'm in the mood for some music. & 1 & user \\ 
         and feelings         &   \hfill  &        \hfill       & \hfill &     \hfill     \\ \hline
        emotional experiences  & 1332 & Can you find me some songs to listen to? & 1 & user \\
         and feelings         &   \hfill  &        \hfill       & \hfill &     \hfill     \\ 
        emotional experiences & 672 & \$15 per head and this is a drama movie. & 11 & system \\ \hline
         and feelings         &   \hfill  &        \hfill       & \hfill &     \hfill     \\ \hline
        emotional experiences & 608 & I am not feeling well, do you know doctor here? & 1 & user \\ 
         and feelings         &   \hfill  &        \hfill       & \hfill &     \hfill     \\ \hline
        emotional experiences & 2205 & I really want to get out and do something. Search things to do in NY please. & 1 & user \\
         and feelings         &   \hfill  &        \hfill       & \hfill &     \hfill     \\ \hline
        sport and leisure & 26 & Thank you. I'd like to find something interesting to do there. I like Games events and I really like Baseball. & 5 & user \\ \hline
        sport and leisure & 1756 & What type of activities are you interested in? & 5 & system \\ \hline
        sport and leisure & 2581 & What are you interested in? Music, Sports, or something else? & 1 & system \\ \hline
        sport and leisure & 3134 & I'm looking for something to do. Can you find me anything involving Sports? & 1 & user \\ \hline
        sport and leisure & 1592 & I'd like a sports event in Oakland please. & 2 & user \\ \hline
        work and school & 84 & That would work. & 3 & user \\ \hline
        work and school & 2066 & Ok, that flight works for me. & 5 & user \\ \hline
        work and school & 2437 & What airport would I have to depart from for this flight? Oh yeah. And tell me if this an economy or business class ticket? & 11 & user \\ \hline
        work and school & 1756 & What type of activities are you interested in? & 5 & system \\ \hline
        work and school & 2205 & I really want to get out and do something. Search things to do in NY please. & 1 & user \\ \hline
    \end{tabular}}
\caption{Example sequences from the \textbf{SGD utterances} topic model}
\end{table*}

\begin{table*}[b]
    \centering
    \begin{tabular}{|l|r|p{.45\linewidth}|r|c|}
    \hline
        ODD-related Keywords & Topic & Sequence Examples & Turn  & Speakers \\ \hline \hline
        personal relationships & 286 & do you like ? & 1 & system \\ \hline
        personal relationships & 593 & What information do you need ? & 1 & system \\ \hline
        personal relationships & 686 & No , I 'm not worried about those things , just as long & 3 & user \\ \hline
        personal relationships & 681 & , do you have something else & 2 & system \\ \hline
        personal relationships & 134 & I would like for & 6 & user \\ \hline
        personal experiences & 487 & that match that description . & 1 & system \\ \hline
        personal experiences & 326 & Do you have a preference as to & 1 & system \\ \hline
        personal experiences & 388 & I do n't have a preference & 7 & user \\ \hline
        personal experiences & 106 & I 'm so bored , can you help me & 3 & user \\ \hline
        personal experiences & 489 & I hope that wo n't discourage you from contacting us again if you need help with & 6 & system \\ \hline
        emotional experiences & 489 & I hope that wo n't discourage you from contacting us again if you need help with & 6 & system \\ 
        and feelings         &   \hfill  &        \hfill       & \hfill &     \hfill     \\ 
        emotional experiences  & 484 & No I would like & 3 & user \\ 
        and feelings         &   \hfill  &        \hfill       & \hfill &     \hfill     \\ 
        emotional experiences  & 650 & , thanks so much for the assistance ! & 7 & user \\ 
        and feelings         &   \hfill  &        \hfill       & \hfill &     \hfill     \\ 
        emotional experiences & 451 & Please contact us anytime . & 6 & system \\ 
        and feelings         &   \hfill  &        \hfill       & \hfill &     \hfill     \\ 
        emotional experiences & 127 & Would that interest you ? & 5 & system \\ 
        and feelings         &   \hfill  &        \hfill       & \hfill &     \hfill     \\ 
        sport and leisure & 285 & I have four attractions & 6 & system \\ \hline
        sport and leisure & 295 & the swimmingpool . & 4 & user \\ \hline
        sport and leisure & 150 & I really hope that the police are able & 3 & system \\ \hline
        sport and leisure & 106 & time ? & 3 & user \\ \hline
        sport and leisure & 497 & I would like to visit a nightclub in the centre . & 2 & user \\ \hline
        work and school & 20 & Would that work for you ? & 5 & system \\ \hline
        work and school & 511 & That would work , & 5 & user \\ \hline
        work and school & 109 & Would you like me to book this for you ? & 4 & system \\ \hline
        work and school & 681 & , do you have something else & 2 & system \\ \hline
        work and school & 250 & Are there any colleges & 1 & user \\ \hline
    \end{tabular}
\caption{Example sequences from the \textbf{filtered MultiWOZ} topic model}
\end{table*}

\begin{table*}[b]
    \begin{center}
    \scalebox{0.79}{%
    \begin{tabular}{|l|r|p{.6\linewidth}|r|c|}
    \hline
        ODD-related Keywords & Topic & Sequence Examples & Turn  & Speakers \\ \hline \hline
        personal relationships & 790 & Thanks. Yes, I am looking for a particular attraction. & 4 & user \\ \hline
        personal relationships & 635 & I'm looking for a college type attraction. & 1 & user \\ \hline
        personal relationships & 79 & What type of attraction is this? & 2 & user \\ \hline
        personal relationships & 8 & That sounds goo. Can you book for 4 people? & 4 & user \\ \hline
        personal relationships & 728 & Hello, I'm looking for a place to go in the north. Could you suggest some places? & 1 & user \\ \hline
        personal experiences & 728 & I'm looking for things to do, specifically a theatre in the north, what kinds of things are there to see like that? & 1 & user \\ \hline
        personal experiences & 1123 & How about any restaurants that serves creative food. & 2 & user \\ \hline
        personal experiences & 790 & Thanks. Yes, I am looking for a particular attraction. & 4 & user \\ \hline
        personal experiences & 340 & There are few options to choose from. How many people are traveling? & 3 & system \\ \hline
        personal experiences & 453 & No, can you recommend me a good one? & 7 & user \\ \hline
        emotional experiences  & 1062 & Feel better soon! Have a better night! & 3 & system \\ 
          and feelings         &   \hfill  &        \hfill       & \hfill &     \hfill     \\ \hline
        emotional experiences & 997 & It was a pleasure being of service. Enjoy your stay in Cambridge and call again if you need any further assistance. Goodbye! & 16 & system \\ 
          and feelings         &   \hfill  &        \hfill       & \hfill &     \hfill     \\ \hline
        emotional experiences & 809 & I am in the mood for Chinese food, please. & 2 & user \\ 
          and feelings         &   \hfill  &        \hfill       & \hfill &     \hfill     \\ \hline
        emotional experiences  & 79 & What type of attraction is this? & 2 & user \\ 
          and feelings         &   \hfill  &        \hfill       & \hfill &     \hfill     \\ \hline
        emotional experiences  & 790 & Thanks. Yes, I am looking for a particular attraction. & 4 & user \\ 
          and feelings         &   \hfill  &        \hfill       & \hfill &     \hfill     \\ \hline
        sport and leisure & 101 & There are no sports in the center. Shall I change type or location for you? & 4 & system \\ \hline
        sport and leisure & 94 & Okay. There are boats, a park, and pools. Would like more information on any of this? & 6 & system \\ \hline
        sport and leisure & 62 & I'm looking for an entertainment venue in the centre of town. Are there any? & 4 & user \\ \hline
        sport and leisure & 982 & I need to know the fee for Parkside Pools please. & 4 & user \\ \hline
        sport and leisure & 95 & The phone number for pizza hut cherry hinton is 01223 323737 and they are located at G4 Cambridge Leisure Park Clifton Way Cherry Hinton. & 4 & system \\ \hline
        work and school & 339 & Thanks! I'm going to hanging out at the college late tonight, could you get me a taxi back to the hotel at 2:45? & 9 & user \\ \hline
        work and school & 200 & I found Old Schools on Trinity Ln, postcode cb21tt. Entrance is free. Would you like to go there? & 1 & system \\ \hline
        work and school & 778 & What day would you like to travel, and at what time do you want to leave? & 1 & system \\ \hline
        work and school & 833 & Is that restaurant moderately priced? & 2 & user \\ \hline
        work and school & 635 & I'm looking for a college type attraction. & 1 & user \\ \hline
    \end{tabular}}
    \end{center}
\caption{Example sequences from the \textbf{MultiWOZ utterances} topic model}
\end{table*}

\end{document}